\documentclass[letterpaper]{article} 
\usepackage{aaai2026}  
\usepackage{times}  
\usepackage{helvet}  
\usepackage{courier}  
\usepackage[hyphens]{url}  
\usepackage{graphicx} 
\usepackage{natbib}  
\usepackage{caption} 
\usepackage{algorithm}
\usepackage{algorithmic}
\usepackage{amsmath}
\usepackage{amsfonts}
\usepackage{color}

\usepackage{booktabs}
\usepackage{multirow}      
\usepackage{colortbl}     
\usepackage{xcolor}       
\usepackage{graphicx}      
\usepackage{float}        
\usepackage{makecell}  

\usepackage{newfloat}
\usepackage{listings}
\DeclareCaptionStyle{ruled}{labelfont=normalfont,labelsep=colon,strut=off} 
\floatstyle{ruled}
\newfloat{listing}{tb}{lst}{}
\floatname{listing}{Listing}
\nocopyright

\title{LoRA in LoRA: Towards Parameter-Efficient Architecture Expansion \\ for Continual Visual Instruction Tuning}
\author{
    Chang Che\textsuperscript{\rm 1}, 
    Ziqi Wang\textsuperscript{\rm 1}, 
    Pengwan Yang\textsuperscript{\rm 2}\thanks{Corresponding authors: yangpengwan2016@gmail.com, zeng\\lin.shi@hfut.edu.cn}, 
    Cheems Wang\textsuperscript{\rm 3}, 
    Hui Ma\textsuperscript{\rm 1}, 
    Zenglin Shi\textsuperscript{\rm 1}\footnotemark[1]
}
\affiliations{
    \textsuperscript{\rm 1}Hefei University of Technology, 
    \textsuperscript{\rm 2}University of Amsterdam, 
    \textsuperscript{\rm 3}Tsinghua University
}

\begin{document}
\maketitle

\begin{abstract}

Continual Visual Instruction Tuning (CVIT) enables Multimodal Large Language Models (MLLMs) to incrementally learn new tasks over time. However, this process is challenged by catastrophic forgetting, where performance on previously learned tasks deteriorates as the model adapts to new ones. A common approach to mitigate forgetting is architecture expansion, which introduces task-specific modules to prevent interference. Yet, existing methods often expand entire layers for each task, leading to significant parameter overhead and poor scalability.
To overcome these issues, we introduce LoRA in LoRA (LiLoRA), a highly efficient architecture expansion method tailored for CVIT in MLLMs. LiLoRA shares the LoRA matrix $A$ across tasks to reduce redundancy, applies an additional low-rank decomposition to matrix $B$ to minimize task-specific parameters, and incorporates a cosine-regularized stability loss to preserve consistency in shared representations over time.
Extensive experiments on a diverse CVIT benchmark show that LiLoRA consistently achieves superior performance in sequential task learning while significantly improving parameter efficiency compared to existing approaches. The code is available at \url{https://github.com/chanceche/LiLoRA}.
\end{abstract}


\section{Introduction}

Multimodal Large Language Models (MLLMs) \cite{bai2025qwen25vl, liu2024improved,zhu2023minigpt} represent a significant advancement over traditional Large Language Models (LLMs) \cite{team2024gemini, touvron2023llama, kaddour2023challenges}, enabling the handling of complex vision-language tasks such as visual question answering (VQA) \cite{chen2024mllm, lee2024visual}, image captioning \cite{awadalla2023openflamingo, liu2024visual}, and visual reasoning \cite{huang2023language, wang2024exploring}. These models are typically trained using a multi-stage pipeline \cite{zhu2023minigpt, liu2024visual, wang2024qwen2vl}, where pretraining on large-scale image-text pairs is followed by visual instruction tuning, aligning model outputs with human intent and improving performance on downstream multimodal tasks.

Visual instruction tuning is commonly performed in a static multi-task setting \cite{dai2023instructblip, liu2024visual}, where all tasks are learned simultaneously using a unified instruction-based format. However, real-world applications increasingly demand that MLLMs continually acquire new capabilities without retraining from scratch. This has led to growing interest in Continual Visual Instruction Tuning (CVIT) \cite{zqwang2024smolora, chen2024coin, he2023continual}, where models incrementally learn new vision-language tasks over time. A major obstacle in this setting is catastrophic forgetting \cite{zhai2023investigating, he2023continual}, in which newly acquired knowledge disrupts or erases information learned from previous tasks.

Most existing CVIT methods adopt static architectures \cite{chen2024coin, zqwang2024smolora, zhao2025mllmcl}, where the model’s structure remains fixed and task-specific routing is used to control parameter sharing. These approaches often incorporate Mixture-of-Experts (MoE) \cite{lepikhin2020gshard, fedus2022switch} modules to reduce interference, but struggle to scale as the number of diverse or unrelated tasks grows. Fixed capacity leads to increased competition among tasks, reducing performance and limiting long-term learning.
To address these limitations, we explore dynamic architecture expansion, a strategy widely used in general continual learning (CL) that introduces task-specific modules as new tasks arrive. While this method offers isolation between tasks, existing CVIT approaches that adopt it \cite{he2023continual} often do so by expanding entire layers of the backbone per task, an approach that quickly becomes inefficient due to significant parameter redundancy and poor scalability in large-scale scenarios.

In this paper, we propose LoRA in LoRA (LiLoRA), a lightweight and scalable architecture expansion method tailored for CVIT in MLLMs. Instead of expanding full layers, LiLoRA builds on Low-Rank Adaptation (LoRA) \cite{hu2021lora}, which expresses fine-tuned updates as a product of two low-rank matrices $A$ and $B$. Through empirical analysis, we observe that the matrix $A$ often converges to similar structures across different tasks. Based on this insight, LiLoRA shares matrix $A$ across all tasks and restricts task-specific adaptation solely to matrix $B$, significantly reducing redundancy.
To further improve parameter efficiency, LiLoRA applies an additional low-rank decomposition to the task-specific matrix $B$, factorizing it into a set of shared basis matrices and task-specific low-rank matrices. This design allows each task to retain flexibility while keeping the overall parameter growth minimal. However, as learning progresses, the shared basis may drift, causing misalignment with previously learned task-specific representations. To counter this, we introduce a cosine-regularized basis stability loss, which penalizes updates to the shared basis based on cosine similarity with prior states, encouraging stability and knowledge retention.

Our contributions are summarized as follows:

\begin{itemize}
\setlength{\itemsep}{0pt}
\item We propose LiLoRA, a parameter-efficient architecture expansion method for CVIT that shares LoRA components across tasks while preserving adaptability through task-specific low-rank decomposition.
\item We introduce a cosine-regularized basis stability loss, which constrains changes to the shared basis and helps retain knowledge over time.
\item We perform extensive experiments on the CVIT Benchmark, showing that LiLoRA achieves state-of-the-art performance while maintaining superior parameter efficiency compared to existing methods.
\end{itemize}

\section{Related Work}

\subsection{CVIT for MLLMs}
 To mitigate the catastrophe forgetting of CVIT, a wide range of approaches have been proposed. CoIN \cite{chen2024coin} applied the token-wise MoE \cite{liu2023moelora, dou2023loramoe} to selectively activate expert weights for different tokens. CL-MoE 
Continual LLaVA \cite{cao2024continual} proposed a novel dual-embedding mechanism combined with selective LoRA modules to mitigate forgetting. LLaCA \cite{qiao2024llaca} designed a gradient-guided exponential moving average strategy to adapt model weights. Fwd-Prompt \cite{zheng2024fwdprompt} leveraged prompt tuning with residual projection to mitigate gradient interference.
MR-LoRA \cite{zhao2025mllmcl} proposed a simple method with domain-specific low-rank tuning and pretrained model-based parameter selection. SMoLoRA \cite{zqwang2024smolora} introduced a separable mixture of low-rank adaptations to address dual forgetting. Although these approaches have demonstrated effectiveness, their static architecture face difficulties in coping with large-scale scenarios. In this paper, we focus on architecture expansion that dynamically adds new parameters to accommodate new tasks.

\subsection{Architecture Expansion for CL}
Architecture expansion is an effective strategy to mitigate catastrophic forgetting during CL. Existing methods \cite{yan2021der,kim2022multihead,douillard2022dytox,xie2024class} typically extend the model with additional modules for each task and applies these task-specific wights to learn new tasks. DER \cite{yan2021der} added task-specific tokens to achieve task-specialized embeddings through a new task-attention layer.
DyTox \cite{douillard2022dytox} introduced new learnable feature extractors with the arrival of new classes to incorporate additional feature dimensions.
MORE \cite{kim2022multihead} and BNCIL \cite{xie2024class} adopted multi-head classification strategies specialized for different classification tasks. Although these methods can preserve previous knowledge from new tasks during CL, when new tasks differ in type from previously seen ones, which is often the case in CVIT, these class-incremental
strategies fail to adapt effectively. In this paper, we focus on efficient  architecture expansion tailored for CVIT.





\begin{figure}
    \includegraphics[width=\linewidth]{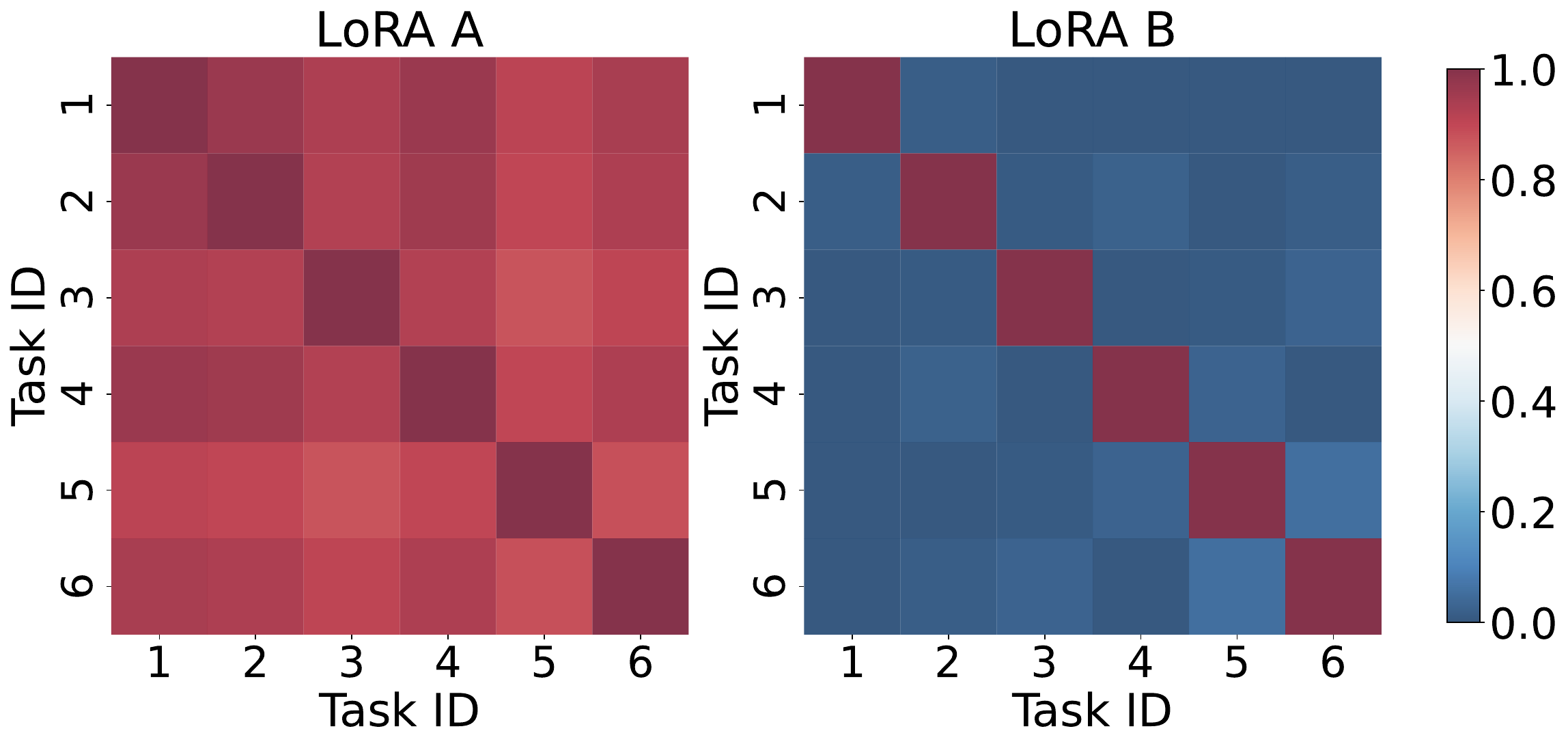}
    \caption{Heatmaps of CKA similarity for LoRA matrices in the linear layers learned by DirLoRA across different tasks. The matrices $A$ exhibit high similarity across tasks, while matrices $B$ show low similarity.}
    \label{fig_heatmap}
\end{figure}

\begin{figure*}
    \centering
    \includegraphics[width=\linewidth]{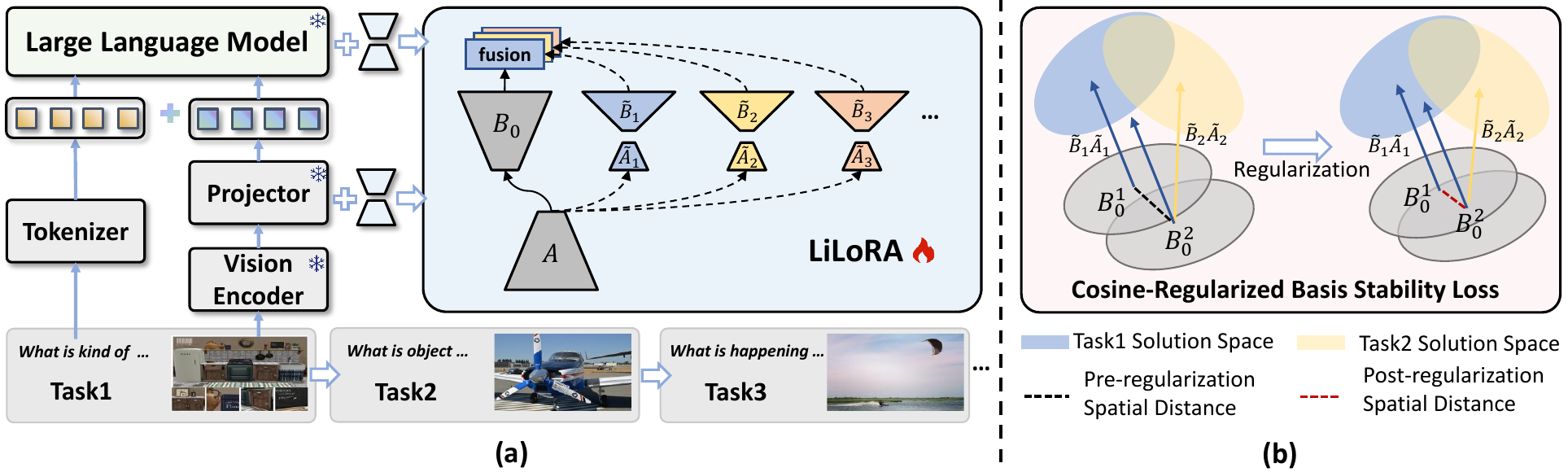}
    \caption{LiLoRA Framework. (a) LiLoRA is an efficient architecture expansion method tailored for CVIT, which can freezes the pretrained model weights and injects lightweight tarinable parameters into existing MLLMs. Specifically, LiLoRA is  initialized with a piar of shared basis matrices and dynamically inserts task-specific low-rank matrices during CVIT. (b) Example of regularization constraining shared basis updates. The gray region represents the shared basis parameter space. The arrows indicate the task-specific residual weight shifts in LiLoRA. When task2 arrives, if the direction of its residual shift exhibits a large angular deviation (i.e., low cosine similarity) from task1, $\mathcal{L}_{reg}$ penalizes large updates to the shared basis, thereby preserving the parameter representations learned from task1.}
    \label{fig_LiLoRA}
\end{figure*}
\section{Methodology}

In CVIT setting, the model is presented with a stream of tasks $\mathcal{T} = \{\tau_1, \tau_2, \dots, \tau_n\}$, where each task $\tau_t$ is associated with a dataset $\mathcal{D}_t = \{(X^{\text{ins}}, X^{\text{inputs}}, X^{\text{gt}})\}$, consisting of textual instructions, visual and textual inputs, and ground-truth response. A fundamental challenge in CVIT is catastrophic forgetting: performance on previously learned tasks degrades as the model updates its parameters to learn new ones. This occurs because shared parameters are overwritten, optimizing them for new tasks at the expense of older ones. To mitigate this, we explore the idea of LoRA-based parameter expansion for CVIT. LoRA has been widely adopted to enable parameter-efficient fine-tuning by introducing a pair of trainable low-rank matrices $B \in \mathbb{R}^{d \times r}$ and $A \in \mathbb{R}^{r \times k}$, with $r \ll \min(d, k)$, into linear layers of the model. Preserving the pretrained weight $W_0 \in \mathbb{R}^{d \times k}$ to be frozen, while $BA$ as the residual weights for adaptation: 
\begin{equation}
W^{'} = W_0 + \Delta  W = W_0 + BA,
\end{equation}
where matrix $B$ is initialized to zeros, while $A$ is drawn from a standard Gaussian distribution. 

A straightforward way to apply the idea of LoRA-based parameter expansion in the CVIT context is Direct LoRA Expansion (DirLoRA), which assigns an independent LoRA module to each task. For a task $\tau_i$, the weight update can be expressed as:
\begin{equation}
  \Delta W_i = B_i A_i,
\end{equation}
where $B_i\in \mathbb{R}^{d \times r}$, $A_i\in \mathbb{R}^{r \times k}$ denote task-specific matrices. While DirLoRA effectively prevents task interference and mitigates forgetting, it introduces substantial parameter overhead, scaling linearly with the number of tasks, which results in inefficient use of model capacity. 

To overcome these limitations, we propose LiLoRA, a more efficient LoRA-based architecture expansion strategy. LiLoRA introduces several key innovations: a shared matrix A across all tasks, a low-rank decomposition of matrix B to further reduce task-specific parameters, and a cosine-regularized stability loss to maintain alignment of the shared components over time. Together, these components enable LiLoRA to preserve performance across tasks while significantly improving parameter efficiency.



\subsection{LiLoRA}


\subsubsection{Task-Invariant Matrix $A$ Sharing.}
To balance parameter efficiency with knowledge retention across sequential tasks, we explore a more effective expansion strategy by investigating the feature representations captured by low-rank matrices. Specifically, we conduct a centered kernel alignment~(CKA) \cite{kornblith2019similarity} similarity analysis on the LoRA matrices learned by DirLoRA. As shown in Fig.~\ref{fig_heatmap}, the matrices $A$ learned across tasks exhibit high similarity, suggesting redundant learning. Motivated by this observation, we propose a shared module design in LiLoRA by reusing the matrix $A$ and limiting task-specific adaptation to matrix $B$. Specifically, we adopt a shared matrix $A \in \mathbb{R}^{r \times k}$ across tasks, while each task $\tau_{i}$ retains its own task-specific matrices $B_i\in \mathbb{R}^{d \times r}$. The weight update for task $\tau_i$ then expressed as:
\begin{equation}
    \Delta  W_i = B_iA.
\end{equation}

This design substantially reduces parameter growth while retaining task-level adaptation capacity.

\subsubsection{Task-Specific Matrix $B$ Decomposition.}
Although the matrix $B$ exhibits lower cross-task similarity, it can be further decomposed into a shared basis and task-specific residuals to improve parameter efficiency. Instead of updating the entire matrix $B$ for each task, we apply task-specific expansion only to its residual component.
For each task $\tau_i$, we introduce a pair of low-rank task-specific matrices $\tilde{B}_i\in \mathbb{R}^{d \times \tilde{r}}$ and $\tilde{A}_i\in \mathbb{R}^{\tilde{r} \times r}$, with $\tilde{r} < r$. Compared to the original matrix $B_i \in \mathbb{R}^{d \times r}$, the product matrix $\tilde{B}_i\tilde{A}_i$ contains significantly fewer parameters, achieving substantial parameter savings while maintaining expressiveness.
Each task's weight is represented as a combination of the shared basis and task-specific matrices:

 \begin{equation}
    \Delta  W_i = (B_0 + \tilde{B}_i\tilde{A}_i)A,
\end{equation}
where the shared matrices $B_0$ and $A$  provide a shared basis across tasks, while the task-specific matrices $\tilde{B}_i$ and $\tilde{A}_i$ specialize the knowledge for the particular task $\tau_i$. Since the importance of shared versus task-specific knowledge may vary across tasks, we introduce a learnable fusion coefficient $\alpha \in (0,1)$ to balance their contributions. The coefficient is initialized as:
\begin{equation}
\alpha \sim \text{Sigmoid}(\mathcal{N}(0,1)),
\end{equation}
where $\mathcal{N}(0, 1)$ denotes the standard Gaussian distribution, and $\text{Sigmoid}(\cdot)$ ensures $\alpha \in (0, 1)$. During training, $\alpha$ is learned via backpropagation, allowing the model to dynamically balance shared and task-specific knowledge. The updated task-specific weight becomes:
\begin{equation}
    \Delta  W_i = (\alpha B_0 + (1-\alpha)\tilde{B}_i\tilde{A}_i)A,
\label{equ_weight}
\end{equation}
a higher $\alpha$ encourages reliance on shared knowledge, while a lower value promotes task-specific adaptation. This adaptive fusion mechanism allows LiLoRA to flexibly tailor its representation to the specific characteristics of each task.

\subsubsection{Decomposition Basis Regularization.}
There is an issue in task-specific matrix $B$ decomposition during the training of task $\tau_t$: updating the shared basis matrix $B_0$ may interfere with the representations learned for previous tasks $\tau_{1}, \dots, \tau_{t-1}$. Although each task retains fixed task-specific matrices $(\tilde{B}_i, \tilde{A}_i)$, modifications to $B_0$ may affect the composite weights $\Delta W_i$ for earlier tasks, potentially causing forgetting of past knowledge.

To mitigate this issue,  we introduce a cosine-regularized basis stability loss, which constrains the magnitude of updates to $B_0$ based on the similarity between task-specific representations. When the new task-specific matrices exhibit low similarity to the previous tasks, the update to $B_0$ should be restricted to preserve the representations of prior tasks. 

Specifically, upon the arrival of a new task $\tau_t$, we compute the cosine similarity between its task-specific matrix product $\tilde{B}_t \tilde{A}_t$ and that of the immediately preceding task $\tilde{B}_{t-1} \tilde{A}_{t-1}$:
\begin{equation}
sim_t =\cos \left(\tilde{B}_{t} \tilde{A}_{t},\tilde{B}_{t-1}\tilde{A}_{t-1} \right),
\end{equation}
The value $sim_t$ then serves as an importance score, scaling the permissible extent of the $B_0$ update. The cosine-regularized basis stability loss is defined as:
\begin{equation}
\mathcal{L}_{reg}=\left(1-sim_t\right) \cdot \left\|B_{0}^{t}-B_{0}^{t-1}\right\|^{2},
\label{equ_reg}    
\end{equation}
where $B_0^{t-1}$ is the value from the previous task, and $B_0^t$ is the current value during task $\tau_t$. As shown in Fig.~\ref{fig_LiLoRA} (b), this loss penalizes large deviations in $B_0$ when the new task's representation ($\tilde{B}_t \tilde{A}_t$) is dissimilar to the previous one, thereby enhancing the stability of the shared basis $B_0$ within the CVIT framework. The overall training procedure for LiLoRA under CVIT is summarized in Algorithm~\ref{alg_Training}.

\begin{algorithm}[tb]
\caption{Training of LiLoRA}
\label{alg_Training}
\textbf{Input}: Dataset $\mathcal{D} = \{\mathcal{D}_1, \mathcal{D}_2, \dots, \mathcal{D}_t\}$, Pretrained model $M$, fusion coefficient $\alpha$, regularization weight $\lambda$

\textbf{Output}: Shared basis matrices $B_0$ and $A$, Task-specific matrices $\{\tilde{B}_t, \tilde{A}_t\}_{t=1}^T$

\begin{algorithmic}[1]
\STATE Freeze pretrained model $M$
\FOR{each Dataset $\mathcal{D}_t$}
    \IF{$t = 1$}
        \STATE Initialize $B_0$, $A$
    \ENDIF
    \STATE Initialize $\tilde{B}_t$, $\tilde{A}_t$
    \FOR{each batch in $\mathcal{D}_t$}
        \STATE Compute weight $\Delta W_t$ using Eq.~\ref{equ_weight}
        \STATE Compute autoregressive loss $\mathcal{L}_{\text{task}}$ for each task $\mathcal{D}_t$ 
        \IF{$t > 1$}
            \STATE Compute regularization loss $\mathcal{L}_{\text{reg}}$ using Eq.~\ref{equ_reg}
        \ELSE
            \STATE $\mathcal{L}_{\text{reg}} \gets 0$
        \ENDIF
        \STATE Minimize $\mathcal{L}_{\text{task}} + \lambda \mathcal{L}_{\text{reg}}$
        \STATE Update $B_0, A, \tilde{B}_t, \tilde{A}_t$
    \ENDFOR
\ENDFOR
\STATE \textbf{return} $B_0$, $A$, $\{\tilde{B}_t, \tilde{A}_t\}_{t=1}^T$
\end{algorithmic}
\end{algorithm}

\begin{table*}[ht]
\normalsize 
\aboverulesep=0pt
\belowrulesep=0pt
\renewcommand\arraystretch{1.25}
\setlength{\tabcolsep}{1mm}
\centering
{
\begin{tabular}{c|c|cccccc|cccc}
\toprule[1.2pt]
\multirow{2}{*}{} &
\multirow{2}{*}{\textbf{Method}} & 
\multicolumn{6}{c}{\textbf{Accuracy on Each Task}} &
\multicolumn{4}{c}{\textbf{Overall Results}} \\ 
{} &{}  & ScienceQA & TextVQA & Flickr30k & ImageNet & GQA & VQAv2 & \textbf{AP} $\uparrow$ &\textbf{ MAP} $\uparrow$ & \textbf{BWT }$\uparrow$ &\textbf{ MIF} $\uparrow$\\ 
\midrule[1pt]
\multirow{12}{*}{\textbf{\textit{Single-type}}}
& {Zero-shot } & 52.72 & 2.95 & 52.64 & 22.10 & 2.73 & 0.65 & 22.30 & - & -  & 17.84 \\  
&  {DirLoRA$^*$ } &  83.75 &  60.66 &  164.20 &  96.71 &  58.55 &  64.93 &  88.13 &  90.18  &  0.00  &  98.41 \\ \cline{2-12}
& {SeqLoRA} & 55.31 & 50.22 & 33.89 & 22.73 & 50.52 & 64.61 & 46.21 & 57.41 & -48.10  & 78.35 \\ 
& {DoRA} & 51.26 & 46.36 & 36.41 & 28.24 & 45.29 & 56.87 & 44.07 & 65.03 & -31.12  & 78.59 \\ 
& {MoeLoRA} & 55.01 & 48.87 & 32.04 & 22.00 & 50.03 & 63.64 & 45.27 & 56.16 & -48.05  & 79.97 \\
& {C-LoRA} & 57.25  & 38.70  & 56.50  & 25.27  & 42.89  & 54.06  &45.78  & 57.04  & -19.58  & 65.84  \\
& {Replay} & 75.61 & 47.58 & 31.97 & 35.84 & 48.51 & 58.67 & 49.70 & 69.78 & -22.71  & 82.06 \\ 
& {EWC} & 57.04 & 50.02 & 32.96 & 22.85 & 50.16 & 64.54 & 46.26 & 56.19 & -49.71  & 78.90 \\ 
& {EWC+TIR} & 72.22  & 44.78  & 34.54  &25.98  & 46.86  & 58.73  & 47.19  & 67.21  & -25.64  & 81.62 	 \\ 

& {Eproj} & 65.29  & 52.87  & 148.19  & 39.45  & 28.06  & 57.86  &65.29  &73.53  & -14.02  & 89.81 	 \\ 
& {SMoLoRA} & 77.36 & 58.29 & 151.99 & 95.35 & 51.96 & \textbf{65.71} & 83.44 & 84.85 & -3.23 &  97.79 \\  \cline{2-12}
 
 &  \textbf{LiLoRA} &   \textbf{77.88} &  \textbf{58.83 }&   \textbf{152.93} &   \textbf{96.02} &    \textbf{58.28 }&   65.33 &   \textbf{84.88} &   \textbf{87.70} &  \textbf{-3.13}  &  \textbf{98.24} \\ 
\midrule[1pt] \midrule[1pt]
\multirow{12}{*}{\textbf{\textit{Five-type}}}
& {Zero-shot} & 51.85 & 5.11 & 44.05 & 20.34 & 2.37 & 1.16 & 20.81 & - & - & 19.45 \\ 
&   {DirLoRA$^*$ } &  83.85 &  60.51 &  164.66 &  96.71 &  57.93 &  64.90 &  88.09 & 91.49   &  0.00 &  98.31 \\ \cline{2-12}
& {SeqLoRA} & 59.21 & 50.80 & 20.99 & 20.30 & 49.98 & 64.41 & 44.28 & 53.75 & -48.73 & 79.47 \\ 
& {DoRA} & 52.03 & 47.37 & 27.97 & 26.18 & 46.05 & 57.33 & 42.82 & 56.24 & -34.11 & 78.30 \\ 
& {MoeLoRA} & 58.09 & 53.30 & 22.82 & 22.61 & 51.80 & 65.15 & 45.63 & 54.88 & -49.94 & 78.19 \\
& {C-LoRA} & 55.58  & 38.64  &59.05  &22.81  & 40.93  & 51.65  & 44.78  & 52.37  & -18.83  & 70.01  \\ 

& {Replay} & 66.06 & 47.78 & 24.21 & 25.66 & 46.53 & 58.59 & 44.81 & 66.68 & -26.88 & 80.38 \\ 
& {EWC} & 53.60 & 49.07 & 20.38 & 20.48 & 50.11 & 64.63 & 43.10 & 53.94 & -52.47  & 78.18 \\ 
& {EWC+TIR} & 66.94  & 45.76  &29.49  & 21.68  & 46.90  &58.80  & 44.93  & 64.51  & -26.38  & 80.56 	 	 \\ 

& {Eproj} & 63.45  & 53.18  & 151.41  & 20.63  & 45.30  & 57.32  & 65.22  & 72.10  & -14.43  & 89.93  \\ 
& {SMoLoRA} &\textbf{ 80.50} & 58.30 & 146.63 & 94.28 & 52.42 & \textbf{65.96} & 83.02 & 85.05 & -6.50 &  98.12 \\  \cline{2-12}

 
 &\textbf{LiLoRA} & 78.38  &\textbf{59.14}&  \textbf{155.26} &   \textbf{95.82} &   \textbf{56.27}&   64.74  &   \textbf{84.94} &   \textbf{87.43} &  \textbf{-1.94}  &  \textbf{98.16} \\ 
\bottomrule[1.2pt]
\end{tabular}
}
\caption{The evaluation results (\%) for continual visual instruction tuning on the CVIT Benchmark \cite{zqwang2024smolora} after training on the final task. *: The performance of DirLoRA serve as an upper-bound for CVIT. LiLoRA consistently maintains high performance across both Single-type and Five-type settings.}
\label{results_main_IF}
\end{table*}

\begin{table*}[t]
\centering
\normalsize 
\aboverulesep=0pt
\belowrulesep=0pt
\renewcommand\arraystretch{1.25}
\setlength{\tabcolsep}{2.3mm}
{
\begin{tabular}{ccc|cccc|cc}
\toprule
\multicolumn{3}{c}{\textbf{Component}} &
\multicolumn{4}{c}{\textbf{Overall Performance}} &
\multicolumn{2}{c}{\textbf{Efficiency}} \\ 
\midrule
\textbf{Share A} & \textbf{Decompose B}& \textbf{${L}_{reg}$} & \textbf{AP} $\uparrow$ & \textbf{MAP} $\uparrow$ & \textbf{BWT} $\uparrow$ & \textbf{MIF} $\uparrow$ & \textbf{TP } $\downarrow$ & \textbf{EP} $\downarrow$ \\
\midrule
-- & -- & -- &  88.13  & 90.18 & 0.00 & 98.41 & 2143.9 & 357.3 \\
 \midrule
 $\checkmark$ &&& 85.38 & 88.40 & -2.87  & 98.39 & 1,250.6& 178.7 \\
 $\checkmark$ &  $\checkmark$& &73.99 & 83.19 & -16.14 &  93.60 &985.1  &   104.6        \\
$\checkmark$ &  $\checkmark$&  $\checkmark$ &84.88  & 87.70 &  -3.13& 98.24 & 985.1  &   104.6       \\

\bottomrule
\end{tabular}
}
\caption{Ablation study on components of LiLoRA and efficiency analysis under the Single-type setting. The 
\textbf{TP} and \textbf{EP} represent the total and each task-specific expansion parameters, respectively, with values given in MB.}
\label{tab_ablation}

\end{table*}

\section{Experiments}
\subsection{Datasets and Evaluation Metrics}
\subsubsection{Datasets.} The datasets used in our experiments are from the CVIT Benchmark \cite{zqwang2024smolora}, which includes six instruction datasets covering visual question answering (VQA) \cite{chen2024mllm, lee2024visual}, image classification \cite{huang2023language, wang2024exploring}, and image captioning \cite{awadalla2023openflamingo, liu2024visual} tasks. Specifically, the benchmark consists of ScienceQA \cite{lu2022learn}, TextVQA \cite{singh2019towards}, Flickr30k \cite{plummer2015flickr30k}, ImageNet \cite{deng2009imagenet}, GQA \cite{hudson2019gqa}, and VQAv2 \cite{goyal2017making}. 

\subsubsection{AP and MAP.} To evaluate overall performance at each learning stage, we compute Average Performance (AP) and Mean Average Performance (MAP) to assess model performance at each learning stage. Specifically, let $a_{k,j}$ denote the accuracy on the $j$-th task (where $j < k$) after training on the $k$-th task. These metrics are defined as:

\begin{equation}
\mathrm{AP}_{k} = \frac{1}{k} \sum_{j=1}^{k} {a}_{k, j}, \quad
\mathrm{MAP}_{k} = \frac{1}{k} \sum_{i=1}^{k} \mathrm{AP}_{i}.
\label{formula_AP}
\end{equation}

\subsubsection{BWT.} To quantify the degree of forgetting, we employ Backward Transfer~(BWT) \cite{wang2024comprehensive}. It is defined as:
\begin{equation}
\mathrm{BWT}_{k} = \frac{1}{k-1} \sum_{j=1}^{k-1} ({a}_{k, j} - {a}_{j,j}).
\label{formula_BWT}
\end{equation}

\subsubsection{MIF.} We adopt the evaluation metric Mean Instruction Following (MIF) \cite{zqwang2024smolora} to evaluate the model's instruction-following consistency. MIF is defined as:

\begin{equation}
\mathrm{MIF}_{k} = \frac{1}{k} \sum_{j=1}^{k} \left( \frac{1}{n} \sum_{i=1}^{n} \mathcal{B}_j(o_i^j) \right),
\label{formula_MIF}
\end{equation}
where $\mathcal{B}()$ is a binary function that returns 1 if the model output $o_i^j$ satisfies the instruction format of the $j$-th task, and 0 otherwise. $n$ denotes the number of evaluation samples.

\subsection{Baseline Methods} 
 We compare our method with a comprehensive set of baselines to highlight its superior performance. \textbf{SeqLoRA }sequentially fine-tunes the model using a single shared LoRA module across tasks. \textbf{DoRA} \cite{liu2024dora} and \textbf{C-LoRA}\cite{smith2023continual} serve as enhanced variants of LoRA designed for fine-tuning. We also include classical CL approaches such as \textbf{EWC} which constrains updates on parameters' importantacne to previous tasks \cite{kirkpatrick2017overcoming}, and \textbf{Replay} \cite{chaudhry2019tiny}
 stores or generates past samples to replay during new task training. In addition, we evaluate several methods tailored for CVIT, including \textbf{MoeLoRA }\cite{liu2024dora}, \textbf{EWC+TIR}, and \textbf{Eproj }\cite{he2023continual}. Notably,\textbf{ Eproj} is an architecture expansion method by extending projection layers based on task similarity. Furthermore, we compare with \textbf{SMoLoRA }\cite{zqwang2024smolora}, a recent state-of-the-art method for CVIT that introduces a separable mixture of low-rank adaptations to address dual forgetting. To provide performance bounds, we include \textbf{DirLoRA}, which ssigns an independent LoRA module for each task as an upper-bound reference, and \textbf{Zero-shot}, which evaluates the pre-trained model without any fine-tuning as a lower-bound reference.

\subsection{Implementation Details}
We adopt the pre-trained first-stage \textbf{LLaVA-v1.5-7B} \cite{liu2024visual} as the base model, without any instruction tuning. The LiLoRA adapters are inserted into the FeedForward Network (FFN) layers of the LLM, as well as into the projection layer between the LLM and the vision encoder. The rank of the shared matrices $r$ in LiLoRA is initialized to $128$, while task-specific $\tilde{r}$ is set to half of $r$. We employ the Adam optimizer with a learning rate of $2\times10^{-5}$ and a batch size of $64$. All tasks are trained for only one epoch.

\begin{table}[t]
\centering
\normalsize 
\aboverulesep=0pt
\belowrulesep=0pt
\setlength{\tabcolsep}{8pt}
\renewcommand\arraystretch{1.25}
\setlength{\tabcolsep}{2.3mm}
{
\begin{tabular}{c|c|cccc}
\toprule
\textbf{$r$} & \textbf{$\tilde{r}$} & \textbf{AP} $\uparrow$ & \textbf{MAP} $\uparrow$ & \textbf{BWT} $\uparrow$ & \textbf{MIF} $\uparrow$ \\
\midrule
\multirow{3}{*}{128} 
  & 64  ($r/2$) & 84.88 & 87.70 & -3.13 & 98.24     \\
  & 32  ($r/4$) &  83.87   &      86.70     &  -2.15      &  97.82         \\
& 16 ($r/8$)   &   83.83      &  86.42       &  -1.55      &  97.94       \\
\midrule
\multirow{3}{*}{64}  
  & 32  ($r/2$) &  84.31       &  87.47       & -2.28       &   97.93      \\
  & 16  ($r/4$) &  83.67       &   86.76      &  -2.51      &     97.63    \\
    & 8  ($r/8$)   &   83.63      &   86.17      &    -1.67    &     97.97    \\
\bottomrule
\end{tabular}
}
\caption{Further analysis on the rank of shared basis ($r$) and task-specific matrices ($\tilde{r}$).}
\label{tab_rank_ablation}

\end{table}

\subsection{Main Results}
    We evaluate LiLoRA on the CVIT Benchmark under two settings, \textit{Single-type instruction} and \textit{Five-type instruction}. After training on the last task, VQAv2, as shown in Table~\ref{results_main_IF}, LiLoRA consistently outperforms all baseline methods. Compared to the recent state-of-the-art method SMoLoRA under the \textit{Single-type instruction} setting, our approach achieves improvements of +1.44\% in AP, +2.85\% in MAP, and +0.10\% in BWT, respectively. In terms of instruction-following ability, our approach also brings a +0.45\% improvement. In comparison to traditional CL approaches such as EWC and Replay, LiLoRA achieves significant superior performance. Notably, when compared with the upper-bound DirLoRA, LiLoRA shows a competitive performance across all results. Consistently, under the \textit{Five-type instruction} setting, LiLoRA achieves improvements of +1.92\% in AP, +2.58\% in MAP, +4.56\% in BWT and +0.04\% in MIF, respectively. These reuslts show that LiLoRA maintains performance compared to other baselines, highlighting its robustness in handling more complex instruction data. 



\subsection{Ablation Study}
In this section, we perform a series of ablation studies to examine the importance of each component in LiLoRA. Specifically, we evaluate the impact of: sharing the LoRA matrix $A$, decomposing the matrix $B$, and incorporating the regularization loss $\mathcal{L}_{reg}$ to stabilize the learned basis. To provide a comprehensive baseline, we also include a DirLoRA setting, where none of these components are applied. As shown in Table \ref{tab_ablation}, solely sharing matrix $A$ achieves competitive performance 85.38\% AP, 88.40\% MAP, -2.87\% BWT and 98.39\% MIF. However, When the matrix $B$ is decomposed without applying $\mathcal{L}_{reg}$, the performance degrades significantly. This degradation can be attributed to the shared basis becoming less aligned with previously learned task-specific matrices over time. With the incorporation of $\mathcal{L}_{reg}$, LiLoRA substantially recovers performance, achieving 84.88\% AP, 87.70\% MAP, -3.13\% BWT, and 98.24\% MIF, close to the results of solely sharing matrix $A$. These results demonstrate the effectiveness of $\mathcal{L}_{\text{reg}}$ in preserving the stability of the shared basis.

For efficiency analysis, we focus on the number of expansion parameters, both in total and task-specific terms. Although DirLoRA achieves strong performance, it suffers from the highest parameter cost, with a total parameter count of 2,143.9MB and 357.3MB each task. In contrast, LiLoRA with all components enabled (shared $A$, decomposed $B$, and $\mathcal{L}_{\text{reg}}$), reduces the total parameters count to 985.1MB and each task parameters to 104.6MB achieving a substantial reduction 54\% in the total parameters and greater savings 70\% in each task overhead compared to DirLoRA. Furthermore, during inference, LiLoRA can be fully merged into the pretrained weights, introducing no extra computational overhead. These results demonstrate that our approach maintains competitive performance while significantly reducing the cost of expansion in CVIT.

\begin{table}[t]
\centering
\normalsize 
\aboverulesep=0pt
\belowrulesep=0pt
\setlength{\tabcolsep}{8pt}
\renewcommand\arraystretch{1.25}
\setlength{\tabcolsep}{2mm}{
\begin{tabular}{c|cccc}
\toprule
{} & \textbf{AP} $\uparrow$ & \textbf{MAP} $\uparrow$ & \textbf{BWT} $\uparrow$ & \textbf{MIF} $\uparrow$ \\
\midrule
$\alpha=1$            & 46.21  & 57.41  &-48.10& 78.35   \\
$\alpha=0$      & 47.01   & 66.73  & -13.45 & 75.53  \\
$\alpha=0.5$      & 81.84  & 87.17  & -6.42 & 97.22  \\
Learnable $\alpha$ (ours)  & \textbf{84.88} & \textbf{87.70} & \textbf{-3.13} & \textbf{98.24} \\
\bottomrule
\end{tabular}
}
\caption{Further analysis on the hyperparameter $\alpha$.}
\label{tab_alpha_ablation}
\end{table}

\begin{figure}
    \includegraphics[width=\linewidth]{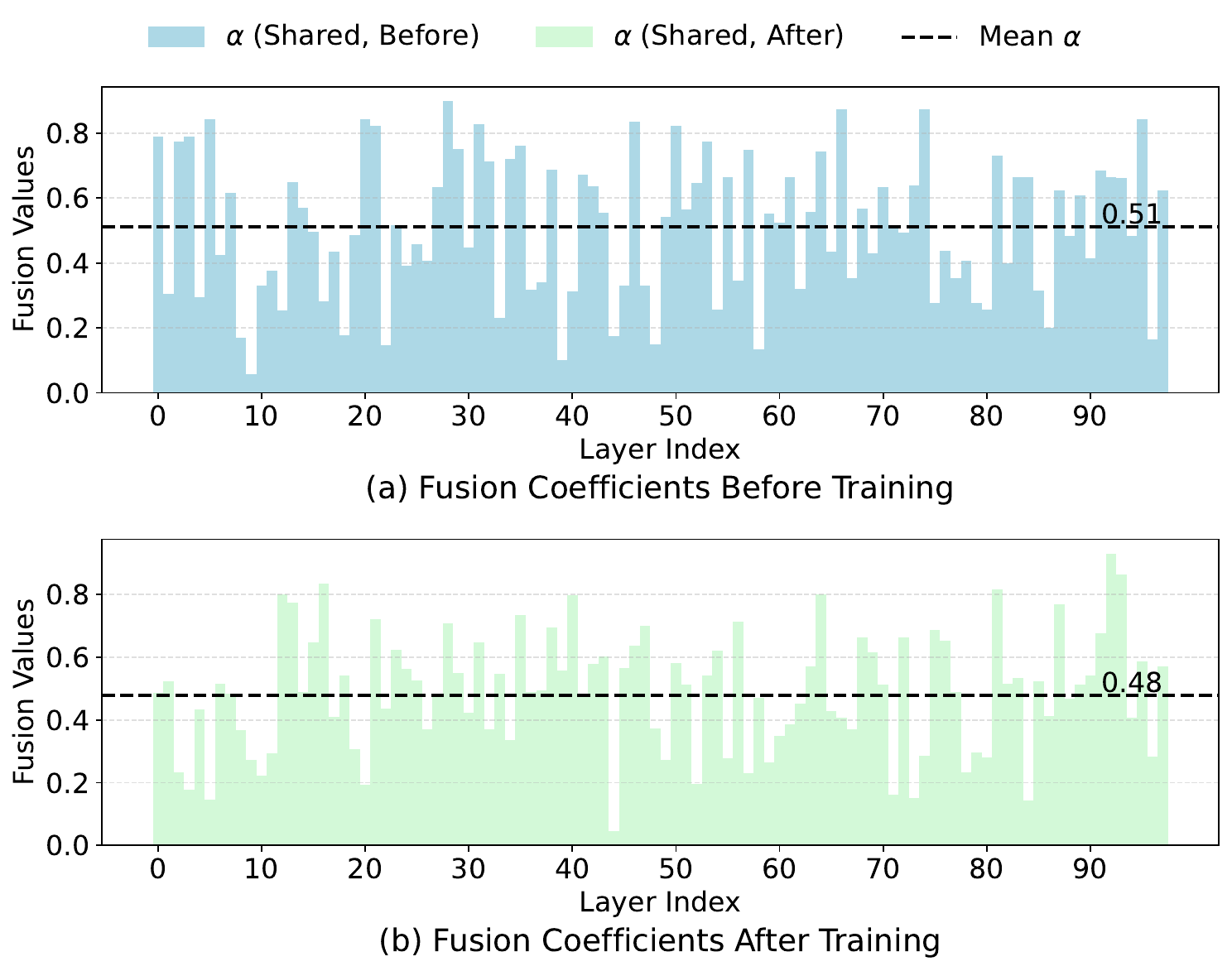}
    \caption{The vlaue of fusion coefficients before and after training on the ScienceQA dataset. (a) shows the initial distribution of fusion values at each layer, while (b) presents the updated values after training. }
    \label{fig_fusion}
\end{figure}





\begin{figure*}[t]
    \centering
    \includegraphics[width=\linewidth]{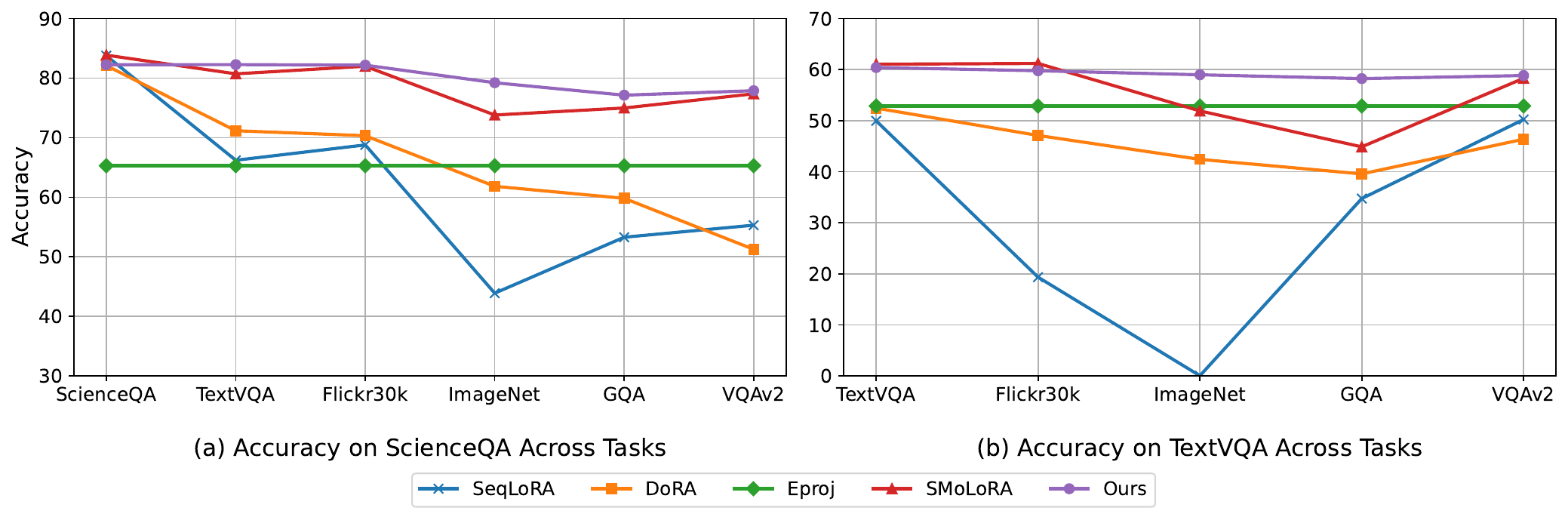}
    \caption{Accuracy (\%) variations curves of ScienceQA and TextVQA during CVIT across sequential tasks. Our method consistently outperforms other methods.}
    \label{fig_cl_acc}
\end{figure*}

\subsection{Further Analysis}
\subsubsection{Effect of the Rank in LiLoRA.}  
To further investigate the impact of hyperparameters in LiLoRA, we explore how the rank of the shared basis ($r$) and the task-specific matrices ($\tilde{r}$) influence the overall performance. As shown in Table~\ref{tab_rank_ablation}, we conduct experiments under the \textit{Single-type instruction} setting with two values of $r$ (64 and 128), and for each, we evaluate $\tilde{r}$ across $\{ r/2, r/4,r/8\}$. The results show that LiLoRA maintains consistently strong performance across a broad range of configurations, even when $\tilde{r}$ is reduced to as low as $r/8$. Although a higher value of $r$ leads to modest gains in certain settings, the performance gap across different $\tilde{r}$ values remains minor, which highlights the robustness of LiLoRA to variations in rank design.
These results demonstrate the flexibility of LiLoRA in selecting rank configurations without significant performance degradation, making it well-suited for real-world scenarios with computational or memory constraints.

\subsubsection{Effect of the Hyperparameter $\alpha$.}
To examine the impact of the hyperparameter $\alpha$ in LiLoRA, we conduct ablation experiments under four settings. First, we fix $\alpha$ to 1, activating only the shared component. Secondly, we set $\alpha$ to 0, enabling only the task-specific component. The third setting assigns a fixed value of 0.5 to $\alpha$, balancing the contributions of both components equally. Lastly, in our proposed approach, $\alpha$ is a learnable parameter and dynamically optimized during training to adapt to task-specific requirements.
The results in Table~\ref{tab_ablation} show that both ($\alpha=0$ or $\alpha=1$) lead to significantly worse performance, indicating that relying solely on either shared or task-specific components is insufficient. The equal weighting ($\alpha=0.5$) shows better results, demonstrating the benefit of combining shared and task-specific weights. Notably, the learnable $\alpha$ achieves the best performance across all metrics, validating the effectiveness of learnable fusion in LiLoRA. 

Furthermore, Fig.~\ref{fig_fusion} illustrates the distributions of fusion coefficients across different layers for the ScienceQA datasets, both before and after training. The dashed lines indicate the mean of $\alpha$, which decreases after training,  suggesting an increasing reliance on the task-specific components. Notably, the $\alpha$ values vary significantly across layers, indicating that the contributions of shared and task-specific components are not uniform throughout the model. These findings further highlighting that LiLoRA can dynamically adjust its dependence on shared versus task-specific components when handling diverse tasks.

\subsubsection{Stability Across Tasks.} 
To comprehensively evaluate LiLoRA’s performance during CVIT, we analyze the task accuracy trends by plotting the representative accuracy curves for two datasets: ScienceQA and TextVQA. These curves illustrate the model’s performance after training on each sequential task, offering a clear depiction of each method’s ability to retain previously learned knowledge while learning a new task. As shown in Fig.~\ref{fig_cl_acc}, the accuracy curve of LiLoRA consistently remains at the top across the entire training sequence for both datasets. In contrast, other methods exhibit a clear downward trend, indicating performance degradation caused by catastrophic forgetting. These findings further highlights the strong stability of our approach compared to competing methods.
\begin{table}[ht]
\centering
\normalsize 
\aboverulesep=0pt
\belowrulesep=0pt
\setlength{\tabcolsep}{8pt}
\renewcommand\arraystretch{1.2}
\setlength{\tabcolsep}{2.3mm}
{
\begin{tabular}{c|cccc}
\toprule
\textbf{Methods} & \textbf{AP} $\uparrow$ & \textbf{MAP} $\uparrow$ & \textbf{BWT} $\uparrow$ & \textbf{MIF} $\uparrow$ \\
\midrule
DirLoRA  & 67.55    & 74.14        & 0.00      &  93.81      \\
\midrule
 SeqLoRA &45.41  &53.39  &-6.42 &   70.91   \\
Eproj     &  58.87     &   61.21     &    -2.47 &  90.12\\ 
 LiLoRA & \textbf{64.63}  & \textbf{68.30}   &\textbf{-0.94}    &\textbf{92.94}    \\
\bottomrule
\end{tabular}
}
\caption{The evaluation results (\%) on \textbf{Qwen2-VL-2B}.}
\label{tab_qwen}
\end{table}
\subsubsection{Cross-Model Generalizability.} To validate the generalization ability of our approach across various MLLMs, we further evaluate the performance of LiLoRA on Qwen2-VL-2B \cite{wang2024qwen2vl} under the \textit{Single-type instruction} setting, comparing it with DirLoRA, SeqLoRA, and Eproj. As shown in Table \ref{tab_qwen}, although Qwen2-VL exhibits a lower degree of forgetting compared to LLaVA, our method still achieves overall performance improvements over SeqLoRA and Eproj, approaching the performance of DirLoRA. These results highlight the robustness of LiLoRA in enhancing performance across diverse models.

\section{Conclusion}
In this paper, we present LiLoRA, a novel and efficient architecture expansion method for CVIT in MLLMs. Motivated by our observation that LoRA matrices $A$ tend to converge to similar representations across tasks, we propose to share matrix $A$ globally and restrict task-specific adaptation solely to matrix $B$, significantly reducing redundancy. To further minimize the parameter footprint, we decompose matrix $B$ into a shared basis and more small task-specific low-rank matrices. To stabilize the shared basis during CVIT, we introduce a cosine-regularized basis stability loss, which helps maintain alignment with previously learned components and mitigates representational drift.
Extensive experiments on the CVIT benchmark demonstrate that LiLoRA not only achieves strong performance across sequential tasks but also offers substantial improvements in parameter efficiency over existing approaches.

\bibliography{aaai2026}

@article{hu2021lora,
  title={Lora: Low-rank adaptation of large language models},
  author={Hu, Edward J and Shen, Yelong and Wallis, Phillip and Allen-Zhu, Zeyuan and Li, Yuanzhi and Wang, Shean and Wang, Lu and Chen, Weizhu},
  journal={arXiv preprint arXiv:2106.09685},
  year={2021}
}

@article{chen2024coin,
  title={CoIN: A Benchmark of Continual Instruction tuNing for Multimodel Large Language Model},
  author={Chen, Cheng and Zhu, Junchen and Luo, Xu and Shen, Hengtao and Gao, Lianli and Song, Jingkuan},
  journal={arXiv preprint arXiv:2403.08350},
  year={2024}
}

@article{zhu2023minigpt,
  title={Minigpt-4: Enhancing vision-language understanding with advanced large language models},
  author={Zhu, Deyao and Chen, Jun and Shen, Xiaoqian and Li, Xiang and Elhoseiny, Mohamed},
  journal={arXiv preprint arXiv:2304.10592},
  year={2023}
}

@inproceedings{liu2024visual,
  author    = {Liu, Haotian and Li, Chunyuan and Wu, Qingyang and Lee, Yong Jae},
  title     = {Visual instruction tuning},
  booktitle = {NeurIPS},
  pages     = {34892--34916},
  year      = {2023}
}

@article{touvron2023llama,
  title={Llama: Open and efficient foundation language models},
  author={Touvron, Hugo and Lavril, Thibaut and Izacard, Gautier and Martinet, Xavier and Lachaux, Marie-Anne and Lacroix, Timoth{\'e}e and Rozi{\`e}re, Baptiste and Goyal, Naman and Hambro, Eric and Azhar, Faisal and others},
  journal={arXiv preprint arXiv:2302.13971},
  year={2023}
}

@article{kaddour2023challenges,
  title={Challenges and applications of large language models},
  author={Kaddour, Jean and Harris, Joshua and Mozes, Maximilian and Bradley, Herbie and Raileanu, Roberta and McHardy, Robert},
  journal={arXiv preprint arXiv:2307.10169},
  year={2023}
}

@article{liu2024dora,
  title={Dora: Weight-decomposed low-rank adaptation},
  author={Liu, Shih-Yang and Wang, Chien-Yi and Yin, Hongxu and Molchanov, Pavlo and Wang, Yu-Chiang Frank and Cheng, Kwang-Ting and Chen, Min-Hung},
  journal={arXiv preprint arXiv:2402.09353},
  year={2024}
}

@article{wang2024comprehensive,
  title={A comprehensive survey of continual learning: Theory, method and application},
  author={Wang, Liyuan and Zhang, Xingxing and Su, Hang and Zhu, Jun},
  journal={IEEE transactions on pattern analysis and machine intelligence},
  volume={46},
  number={8},
  pages={5362--5383},
  year={2024},
  publisher={IEEE}
}

@article{he2023continual,
  title={Continual instruction tuning for large multimodal models},
  author={He, Jinghan and Guo, Haiyun and Tang, Ming and Wang, Jinqiao},
  journal={arXiv preprint arXiv:2311.16206},
  year={2023}
}

@article{dai2023instructblip,
  title={Instructblip: Towards general-purpose vision-language models with instruction tuning. arxiv 2023},
  author={Dai, Wenliang and Li, Junnan and Li, D and Tiong, AMH and Zhao, J and Wang, W and Li, B and Fung, P and Hoi, S},
  journal={arXiv preprint arXiv:2305.06500},
  volume={2},
  year={2023}
}

@inproceedings{lu2022learn,
  author    = {Lu, Pan and Mishra, Swaroop and Xia, Tony and Qiu, Liang and Chang, Kai-Wei and Zhu, Song-Chun and Tafjord, Oyvind and Clark, Peter and Kalyan, Ashwin},
  title     = {Learn to Explain: Multimodal Reasoning via Thought Chains for Science Question Answering},
  booktitle = {NeurIPS},
  year      = {2022},
  pages     = {2507--2521},

}

@inproceedings{singh2019towards,
  title={Towards vqa models that can read},
  author={Singh, Amanpreet and Natarajan, Vivek and Shah, Meet and Jiang, Yu and Chen, Xinlei and Batra, Dhruv and Parikh, Devi and Rohrbach, Marcus},
  booktitle={Proceedings of the IEEE/CVF Conference on Computer Vision and Pattern Recognition},
  pages={8317--8326},
  year={2019}
}

@inproceedings{plummer2015flickr30k,
  title={Flickr30k entities: Collecting region-to-phrase correspondences for richer image-to-sentence models},
  author={Plummer, Bryan A and Wang, Liwei and Cervantes, Chris M and Caicedo, Juan C and Hockenmaier, Julia and Lazebnik, Svetlana},
  booktitle={Proceedings of the IEEE/CVF International Conference on Computer Vision},
  pages={2641--2649},
  year={2015}
}

@inproceedings{deng2009imagenet,
  title={Imagenet: A large-scale hierarchical image database},
  author={Deng, Jia and Dong, Wei and Socher, Richard and Li, Li-Jia and Li, Kai and Fei-Fei, Li},
  booktitle={2009 IEEE conference on computer vision and pattern recognition},
  pages={248--255},
  year={2009}
}

@inproceedings{hudson2019gqa,
  title={Gqa: A new dataset for real-world visual reasoning and compositional question answering},
  author={Hudson, Drew A and Manning, Christopher D},
  booktitle={Proceedings of the IEEE/CVF Conference on Computer Vision and Pattern Recognition},
  pages={6700--6709},
  year={2019}
}

@inproceedings{goyal2017making,
  author    = {Yash Goyal and Tejas Khot and Douglas Summers-Stay and Dhruv Batra and Devi Parikh},
  title     = {Making the v in vqa matter: Elevating the role of image understanding in visual question answering},
  booktitle = {Proceedings of the IEEE Conference on Computer Vision and Pattern Recognition (CVPR)},
  pages     = {6904--6913},
  year      = {2017},

}

@article{kirkpatrick2017overcoming,
  title={Overcoming catastrophic forgetting in neural networks},
  author={Kirkpatrick, James and Pascanu, Razvan and Rabinowitz, Neil and Veness, Joel and Desjardins, Guillaume and Rusu, Andrei A and Milan, Kieran and Quan, John and Ramalho, Tiago and Grabska-Barwinska, Agnieszka and others},
  journal={Proceedings of the national academy of sciences},
  volume={114},
  number={13},
  pages={3521--3526},
  year={2017},
  publisher={National Acad Sciences}
}

@article{zhai2023investigating,
  title={Investigating the catastrophic forgetting in multimodal large language models},
  author={Zhai, Yuexiang and Tong, Shengbang and Li, Xiao and Cai, Mu and Qu, Qing and Lee, Yong Jae and Ma, Yi},
  journal={arXiv preprint arXiv:2309.10313},
  year={2023}
}

@article{liu2023moelora,
  title={Moelora: An moe-based parameter efficient fine-tuning method for multi-task medical applications},
  author={Liu, Qidong and Wu, Xian and Zhao, Xiangyu and Zhu, Yuanshao and Xu, Derong and Tian, Feng and Zheng, Yefeng},
  journal={arXiv preprint arXiv:2310.18339},
  year={2023}
}

@article{dou2023loramoe,
  title={Loramoe: Revolutionizing mixture of experts for maintaining world knowledge in language model alignment},
  author={Dou, Shihan and Zhou, Enyu and Liu, Yan and Gao, Songyang and Zhao, Jun and Shen, Wei and Zhou, Yuhao and Xi, Zhiheng and Wang, Xiao and Fan, Xiaoran and others},
  journal={arXiv preprint arXiv:2312.09979},
  volume={4},
  number={7},
  year={2023}
}

@article{huang2023language,
  author    = {Shaohan Huang and Li Dong and Wenhui Wang and Yaru Hao and Saksham Singhal and Shuming Ma and Tengchao Lv and Lei Cui and Owais Khan Mohammed and Barun Patra and Qiang Liu and Kriti Aggarwal and Zewen Chi and Johan Bjorck and Vishrav Chaudhary and Subhojit Som and Xia Song and Furu Wei},
  title     = {Language Is Not All You Need: Aligning Perception with Language Models},
  journal   = {arXiv preprint arXiv:2302.14045},
  year      = {2023},

}

@article{wang2024exploring,
  title={Exploring the reasoning abilities of multimodal large language models (mllms): A comprehensive survey on emerging trends in multimodal reasoning},
  author={Wang, Yiqi and Chen, Wentao and Han, Xiaotian and Lin, Xudong and Zhao, Haiteng and Liu, Yongfei and Zhai, Bohan and Yuan, Jianbo and You, Quanzeng and Yang, Hongxia},
  journal={arXiv preprint arXiv:2401.06805},
  year={2024}
}

@article{chen2024mllm,
  title={Mllm-as-a-judge: Assessing multimodal llm-as-a-judge with vision-language benchmark},
  author={Chen, Dongping and Chen, Ruoxi and Zhang, Shilin and Liu, Yinuo and Wang, Yaochen and Zhou, Huichi and Zhang, Qihui and Wan, Yao and Zhou, Pan and Sun, Lichao},
  journal={arXiv preprint arXiv:2402.04788},
  year={2024}
}

@article{lee2024visual,
  title={Visual question answering instruction: Unlocking multimodal large language model to domain-specific visual multitasks},
  author={Lee, Jusung and Cha, Sungguk and Lee, Younghyun and Yang, Cheoljong},
  journal={arXiv preprint arXiv:2402.08360},
  year={2024}
}

@article{awadalla2023openflamingo,
  title={Openflamingo: An open-source framework for training large autoregressive vision-language models},
  author={Awadalla, Anas and Gao, Irena and Gardner, Josh and Hessel, Jack and Hanafy, Yusuf and Zhu, Wanrong and Marathe, Kalyani and Bitton, Yonatan and Gadre, Samir and Sagawa, Shiori and others},
  journal={arXiv preprint arXiv:2308.01390},
  year={2023}
}

@article{lepikhin2020gshard,
  title={Gshard: Scaling giant models with conditional computation and automatic sharding},
  author={Lepikhin, Dmitry and Lee, HyoukJoong and Xu, Yuanzhong and Chen, Dehao and Firat, Orhan and Huang, Yanping and Krikun, Maxim and Shazeer, Noam and Chen, Zhifeng},
  journal={arXiv preprint arXiv:2006.16668},
  year={2020}
}

@article{fedus2022switch,
  title={Switch transformers: Scaling to trillion parameter models with simple and efficient sparsity},
  author={Fedus, William and Zoph, Barret and Shazeer, Noam},
  journal={Journal of Machine Learning Research},
  volume={23},
  number={120},
  pages={1--39},
  year={2022}
}

@article{chaudhry2019tiny,
  author    = {Arslan Chaudhry and Marcus Rohrbach and Mohamed Elhoseiny and Thalaiyasingam Ajanthan and Puneet K. Dokania and Philip H. S. Torr and Marc'Aurelio Ranzato},
  title     = {Continual learning with tiny episodic memories},
  journal   = {arXiv preprint arXiv:1902.10486},
  year      = {2019},
  url       = {https://arxiv.org/abs/1902.10486}
}

@article{smith2023continual,
  title={Continual diffusion: Continual customization of text-to-image diffusion with c-lora},
  author={Smith, James Seale and Hsu, Yen-Chang and Zhang, Lingyu and Hua, Ting and Kira, Zsolt and Shen, Yilin and Jin, Hongxia},
  journal={arXiv preprint arXiv:2304.06027},
  year={2023}
}

@article{zqwang2024smolora,
  title={SMoLoRA: Exploring and Defying Dual Catastrophic Forgetting in Continual Visual Instruction Tuning},
  author={Wang, Ziqi and Che, Chang and Wang, Qi and Li, Yangyang and Shi, Zenglin and Wang, Meng},
  journal={arXiv preprint arXiv:2411.13949},
  year={2024},
}

@article{kim2022multihead,
  title={A Multi-Head Model for Continual Learning via Out-of-Distribution Replay},
  author={Kim, Gyuhak and Ke, Zixuan and Liu, Bing},
  journal={arXiv preprint arXiv:2208.09734},
  year={2022},
}

@inproceedings{douillard2022dytox,
  title     = {{DyTox: Transformers for Continual Learning with Dynamic Token Expansion}},
  author    = {Douillard, Arthur and Rame, Alexandre and Couairon, Guillaume and Cord, Matthieu},
  booktitle = {Proceedings of the IEEE/CVF Conference on Computer Vision and Pattern Recognition },
  year      = {2022},
  pages     = {9285--9295}
}

@inproceedings{yan2021der,
  title     = {{DER: Dynamically Expandable Representation for Class Incremental Learning}},
  author    = {Yan, Shuang and Xie, Jingkuan and He, Xianglong},
  booktitle = {Proceedings of the IEEE/CVF Conference on Computer Vision and Pattern Recognition},
  year      = {2021},
  pages     = {3014--3023}
}

@article{xie2024class,
  title={Class Incremental Learning with Task-Specific Batch Normalization and Out-of-Distribution Detection},
  author={Xie, Xuchen and Qiu, Yiqiao and Lin, Run and Zheng, Weishi and Wang, Ruixuan},
  journal={arXiv preprint arXiv:2411.00430},
  year={2024},
}

@article{kornblith2019similarity,
  title={Similarity of neural network representations revisited},
  author={Kornblith, Simon and Norouzi, Mohammad and Lee, Honglak and Hinton, Geoffrey},
  journal={arXiv preprint arXiv:1905.00414},
  year={2019}
}

@article{cao2024continual,
  title={Continual LLaVA: Continual Instruction Tuning in Large Vision-Language Models},
  author={Cao, Meng and Liu, Yuyang and Liu, Yingfei and Wang, Tiancai and Dong, Jiahua and Ding, Henghui and Zhang, Xiangyu and Reid, Ian and Liang, Xiaodan},
  journal={arXiv preprint arXiv:2411.02564},
  year={2024}
}

@article{qiao2024llaca,
  title={Large Continual Instruction Assistant},
  author={Qiao, Jingyang and Zhang, Zhizhong and Tan, Xin and Qu, Yanyun and Ding, Shouhong and Xie, Yuan},
  journal={arXiv preprint arXiv:2410.10868},
  year={2024}
}

@article{wang2024qwen2vl,
  title={Qwen2-VL: Enhancing Vision-Language Model's Perception of the World at Any Resolution},
  author={Wang, Peng and Bai, Shuai and Tan, Sinan and Wang, Shijie and Fan, Zhihao and Bai, Jinze and Chen, Keqin and Liu, Xuejing and Wang, Jialin and Ge, Wenbin and Fan, Yang and Dang, Kai and Du, Mengfei and Ren, Xuancheng and Men, Rui and Liu, Dayiheng and Zhou, Chang and Zhou, Jingren and Lin, Junyang},
  year={2024},
  journal={arXiv preprint arXiv:2408.15262},

}

@article{zhao2025mllmcl,
  author    = {H. Zhao and F. Zhu and R. Wang and G. Meng and Z. Zhang},
  title     = {MLLM-CL: Continual Learning for Multimodal Large Language Models},
  journal   = {arXiv preprint arXiv:2506.05453},
  year      = {2025}
}

@inproceedings{liu2024improved,
  author    = {H. Liu and C. Li and Y. Li and Y. J. Lee},
  title     = {Improved Baselines with Visual Instruction Tuning},
  booktitle = {Proceedings of the IEEE/CVF Conference on Computer Vision and Pattern Recognition (CVPR)},
  pages     = {26296--26306},
  year      = {2024}
}

@article{bai2025qwen25vl,
  author    = {S. Bai and K. Chen and X. Liu and J. Wang and W. Ge and S. Song and K. Dang and P. Wang and S. Wang and J. Tang and others},
  title     = {Qwen2.5-VL Technical Report},
  journal   = {arXiv preprint arXiv:2502.13923},
  year      = {2025}
}

@article{team2024gemini,
  author    = {G. Team and P. Georgiev and V. I. Lei and R. Burnell and L. Bai and A. Gulati and G. Tanzer and D. Vincent and Z. Pan and S. Wang and others},
  title     = {Gemini 1.5: Unlocking Multimodal Understanding Across Millions of Tokens of Context},
  journal   = {arXiv preprint arXiv:2403.05530},
  year      = {2024}
}

@article{zheng2024fwdprompt,
  title={Beyond Anti-Forgetting: Multimodal Continual Instruction Tuning with Positive Forward Transfer},
  author={Zheng, Jiayi and Ma, Qian and Liu, Zhen and Wu, Bing and Feng, Huijuan},
  journal={arXiv preprint arXiv:2401.09181},
  year={2024}
}
\end{document}